\documentclass[conference]{IEEEtran}
\IEEEoverridecommandlockouts

\usepackage{cite}
\usepackage{amsmath,amssymb,amsfonts}
\usepackage{multirow}
\usepackage{array}
\usepackage{arydshln}  
\usepackage{colortbl}  
\usepackage{algorithmic}
\usepackage{graphicx}
\usepackage{textcomp}
\usepackage{xcolor}
\usepackage{booktabs}  
\usepackage{makecell}  
\usepackage{threeparttable}  
\usepackage{url}
\usepackage{balance}

\def\BibTeX{{\rm B\kern-.05em{\sc i\kern-.025em b}\kern-.08em
    T\kern-.1667em\lower.7ex\hbox{E}\kern-.125emX}}
\begin{document}

\title{Fine-tuning Large Language Models for Entity Matching}

\author{\IEEEauthorblockN{Aaron Steiner}
\IEEEauthorblockA{\textit{University of Mannheim}\\
Mannheim, Germany \\
aaron.steiner@uni-mannheim.de}
\and
\IEEEauthorblockN{Ralph Peeters}
\IEEEauthorblockA{\textit{University of Mannheim}\\
Mannheim, Germany \\
ralph.peeters@uni-mannheim.de}
\and
\IEEEauthorblockN{Christian Bizer}
\IEEEauthorblockA{\textit{University of Mannheim}\\
Mannheim, Germany \\
christian.bizer@uni-mannheim.de}
}
\maketitle

\begin{abstract}
Generative large language models (LLMs) are a promising alternative to pre-trained language models for entity matching due to their high zero-shot performance and ability to generalize to unseen entities. Existing research on using LLMs for entity matching has focused on prompt engineering and in-context learning. This paper explores the potential of fine-tuning LLMs for entity matching. We analyze fine-tuning along two dimensions: 1) the representation of training examples, where we experiment with adding different types of LLM-generated explanations to the training set, and 2) the selection and generation of training examples using LLMs. In addition to the matching performance on the source dataset, we investigate how fine-tuning affects the model’s ability to generalize to other in-domain datasets as well as across topical domains. Our experiments show that fine-tuning significantly improves the performance of the smaller models while the results for the larger models are mixed. Fine-tuning also improves the generalization to in-domain datasets while hurting cross-domain transfer. We show that adding structured explanations to the training set has a positive impact on the performance of three out of four LLMs, while the proposed example selection and generation methods, only improve the performance of Llama 3.1 8B while decreasing the performance of GPT-4o-mini. 
\end{abstract}


\section{Introduction}
\label{ref:introduction}

\begin{figure}[h]
\centering
\includegraphics[width=1\linewidth]{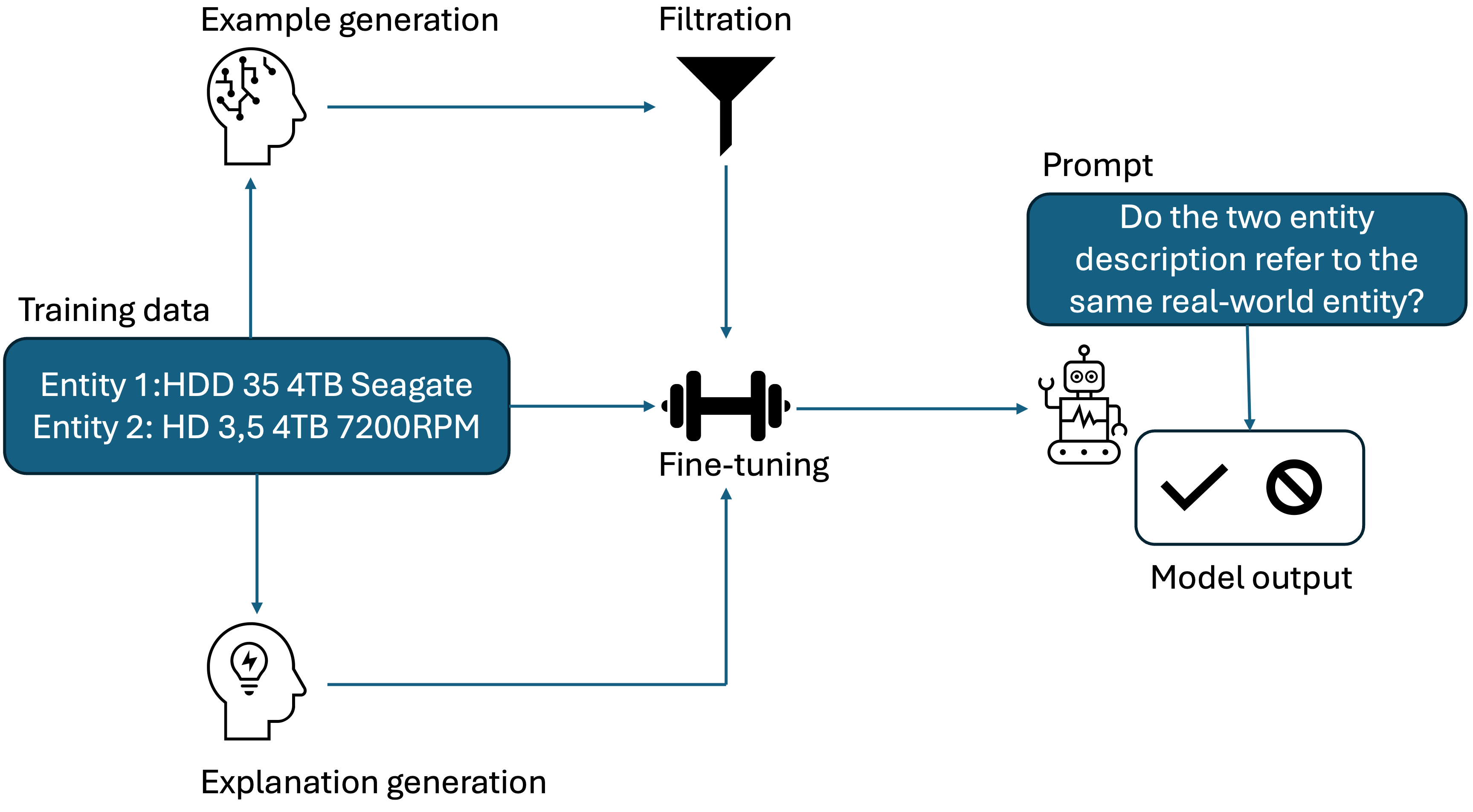}
\caption{Fine-tuning and inference setup including explanation generation, example generation, and filtration.} \label{fig:fine_tuning_setup}
\end{figure}

Entity matching~\cite{BarlaugNeural2021,Christen2012DataMC,elmagarmidDuplicateRecordDetection2007} is the task of identifying entity descriptions in different data sources that refer to the same real-world entity. Entity matching is a central step in data integration pipelines~\cite{christophides_end--end_2020}. 
Many state-of-the-art entity matching methods build on pre-trained language models (PLMs)~\cite{vaswaniAttentionAllYou2017}, such as BERT or RoBERTa~\cite{liDeepEntityMatching2020, peetersDualobjectiveFinetuningBERT2021, peeters2023wdc, zeakis2023pre}.
Recent work on using generative large language models (LLMs)~\cite{zhao2023survey}, such as GPT, Llama, Gemini, or Mistral, for entity matching has shown that LLMs have higher zero-shot performance compared to PLMs and generalize better to unseen entities~\cite{narayan2022can,peetersUsingChatGPTEntity2023a,peeters2025entitymatchingusinglarge}.
Most research on using LLMs for entity matching focuses on prompt engineering and in-context learning~\cite{narayan2022can,peetersUsingChatGPTEntity2023a, fanCostEffectiveInContextLearning2023}. Initial papers have started to investigate the potential of improving beyond in-context learning by fine-tuning LLMs for entity matching~\cite{peeters2025entitymatchingusinglarge,zhangJellyfishLargeLanguage2023}.
This paper builds on this work and presents a deeper investigation of the utility of fine-tuning LLMs for entity matching. Besides the effect of fine-tuning on the matching performance, we investigate how fine-tuning influences the model's ability to generalize to different in-domain datasets as well as across topical domains. Our investigation focuses on two dimensions: First, the representation of training examples, where we investigate the effect of augmenting each training example with textual as well as structured explanations about why a pair of entity descriptions matches or not. The second dimension is the selection and generation of examples for the training set that is used for fine-tuning. We experiment with using LLMs to filter irrelevant examples from the training set as well as to generate additional training examples. We further investigate the error-based selection of training examples. Figure~\ref{fig:fine_tuning_setup} gives an overview of the fine-tuning and inference setup. The effects of fine-tuning are compared between open-source (Llama-3.1) and proprietary (GPT-4o) models.

\textbf{Contributions:} This paper makes the following contributions:

\begin{enumerate}
    \item \textbf{Standard Fine-Tuning:} To establish baselines, we fine-tune open source and proprietary LLMs of different sizes using standard fine-tuning techniques. The baselines show that fine-tuning significantly improves the performance of the smaller models while the results for the larger models are mixed. 
    \item \textbf{In-domain and Cross-domain Generalization:} We assess the effectiveness of fine-tuned models in both in-domain and cross-domain settings. The experiments show that fine-tuned models often generalize better to in-domain datasets, while their performance in cross-domain settings falls short of the zero-shot baseline in the majority of cases.
    \item \textbf{Example Representation:} We investigate the impact of augmenting the training set with different types of LLM-generated explanations. Adding structured explanations improves the performance of three out of four fine-tuned models, while also improving in-domain generalization.
    \item \textbf{Example Selection and Generation:} We introduce various approaches for using LLMs to filter the training set and generate new examples. While these methods yield notable gains for Llama~3.1\,8B, they result in performance degradation for GPT-4o-mini.
  \end{enumerate}

\section{Experimental Setup}
\label{sec:experimentalsetup}
This section provides details on the models, benchmark datasets, fine-tuning setup, and the performance evaluation that we use for the experiments. All artifacts are provided in the project repository, meaning that all experiments can be replicated\footnote{https://github.com/wbsg-uni-mannheim/TailorMatch}.

\textbf{Large Language Models:} We compare the following hosted proprietary models and locally-run open-source models:

\begin{enumerate}
\item \textbf{gpt-4o-2024-08-06:} This LLM was released in August 2024 as OpenAI’s flagship model and is used by ChatGPT.
\item \textbf{gpt-4o-mini-2024-07-18:} This hosted LLM, released in July 2024, has lower usage fees compared to GPT-4o. Although the model size is not disclosed, the naming convention indicates that it is smaller than GPT-4o.
\item \textbf{Meta-Llama-3.1-8B-Instruct\footnote{https://huggingface.co/meta-llama/Meta-Llama-3.1-8B-Instruct}:} Released in July 2024, this open-source model from Meta features the same 128k token context length as the GPT models.
\item \textbf{Meta-Llama-3.1-70B-Instruct\footnote{meta-llama/Meta-Llama-3.1-70B-Instruct}:} Also developed by Meta, this model is medium-sized within the Llama series. Further details are available in the accompanying paper by Meta~\cite{grattafiori2024llama}.
\end{enumerate}

We use the OpenAI API\footnote{https://platform.openai.com/docs/guides/batch} and Hugging Face’s Transformers\footnote{https://huggingface.co/docs/transformers/en/index} library to prompt both hosted and local models. In both cases, the LLM temperature is set to zero in order to minimize randomness. We run open-source models on a multi-GPU server equipped with up to eight Nvidia RTX6000 GPUs, using one GPU for LLaMA 8B and four GPUs for the 70B model.

\textbf{Datasets:} We evaluate in-domain and cross-domain generalization using benchmark datasets from two topical domains~\cite{kopckeEvaluationEntityResolution2010b,peeters2023wdc}: products and scholarly works.    
Each dataset has at least 150 matches in its test set (Table~\ref{tab:dataset_stats}) to ensure stable measurement. Second, we prioritize datasets that contain a significant number of challenging corner case record pairs. These corner cases are either matching or non-matching pairs that closely resemble the opposite class due to very similar or dissimilar surface forms~\cite{peeters2023wdc}. Finally, to evaluate cross-domain generalization, we chose datasets from two topical domains. We use the following datasets:

\begin{itemize}
    \item \textbf{WDC Products:} This dataset includes product offers from various categories (e.g., electronics, clothing) originating from numerous online shops. We use the most challenging version, with 80\% corner cases (hard positives and negatives).
    \item \textbf{Abt-Buy}: This benchmark provides examples within similar categories to those found in WDC Products.
    \item \textbf{Walmart-Amazon:} This benchmark includes products from both Walmart and Amazon, with categories comparable to those in WDC Products and Abt-Buy.
    \item \textbf{Amazon-Google:} Unlike the other benchmarks, the Amazon-Google dataset focuses on software products, such as different versions of the Windows operating system and various image/video editing applications. This dataset introduces a different product type for evaluating model performance.
    \item \textbf{DBLP-Scholar:} This benchmark requires matching bibliographic entries between DBLP and Google Scholar, representing a domain distinct from the product datasets.
    \item \textbf{DBLP-ACM:} Similar to DBLP-Scholar, the DBLP-ACM benchmark involves matching bibliographic entries.
\end{itemize}

We use the title attribute of the product datasets for fine-tuning, while we concatenate the author, title, venue, and year attributes of the bibliographic datasets using a semicolon character as a delimiter.



\begin{table}[ht]
    \centering
    \scriptsize
    \caption{Dataset statistics for the training, validation, and test sets.}
    \label{tab:dataset_stats}
    \resizebox{\columnwidth}{!}{%
    \begin{tabular}{l|rr|rr|rr}
        \toprule
        \textbf{Dataset} & \multicolumn{2}{c|}{\textbf{Training Set}} & \multicolumn{2}{c|}{\textbf{Validation Set}} & \multicolumn{2}{c}{\textbf{Test Set}} \\
         & \textbf{\# Pos} & \textbf{\# Neg} & \textbf{\# Pos} & \textbf{\# Neg} & \textbf{\# Pos} & \textbf{\# Neg} \\
        \midrule
        WDC Products (small) & 500 & 2,000 & 500 & 2,000  & 500 & 4,000 \\
        WDC Products (medium) & 1,500 & 4,500 & 500 & 3,000 & 500 & 4,000 \\
        WDC Products (large) & 8,471 & 11,364 & 500 & 4,000 & 500 & 4,000 \\
        Abt-Buy (A-B) & 822 & 6,837 & 206 & 1,710 & 206 & 1,710 \\
        Amazon-Google (A-G) & 933 & 8,234 & 234 & 2,059 & 234 & 2,059 \\
        Walmart-Amazon (W-A) & 769 & 7,424 & 193 & 1,856 & 193 & 1,856 \\
        DBLP-Scholar (D-S) & 4,277 & 18,688 & 1,070 & 4,672 & 1,070 & 4,672 \\
        DBLP-ACM (D-A) & 1,776 & 8,114 & 444 & 2,029 & 444 & 2,029 \\
        \bottomrule
    \end{tabular}%
    }
\end{table}

\textbf{Fine-tuning Setup:} 
We use the default hyperparameter setting recommended by the model providers.
The fine-tuning setup differs for hosted and open-source models. We use the following default hyperparameter settings for the hosted OpenAI models: learning rate multiplier 1.8, batch size 16. 
We use the following fine-tuning setup for the open-source models: We employed Low-Rank Adaptation (LoRA) with an alpha of 16 to allow moderate adjustments to model weights, a dropout rate of 0.1 to prevent overfitting, and a rank (r) of 64 to balance performance and computational efficiency. The learning rate was configured at 2e-4. 
Both open-source and hosted models are trained for 10 epochs, with checkpoints generated after each epoch. Open-source models are validated at each checkpoint using custom evaluation callbacks. In contrast, OpenAI models provided only two intermediate checkpoints plus the final version, limiting the validation process.

\textbf{Evaluation:} Following related work~\cite{BarlaugNeural2021}, we use precision, recall, and F1. All subsequent tables show the F1 performance, while precision and recall results are provided in the accompanying repository.
Model responses in natural language are evaluated using Narayan et al.'s~\cite{narayan2022can} method of parsing responses for ``yes'' or ``no'' decisions.
\section{Standard Fine-Tuning}
\label{sec:standard_fine_tuning}
\begin{figure}[h!]
\centering
\includegraphics[width=\linewidth]{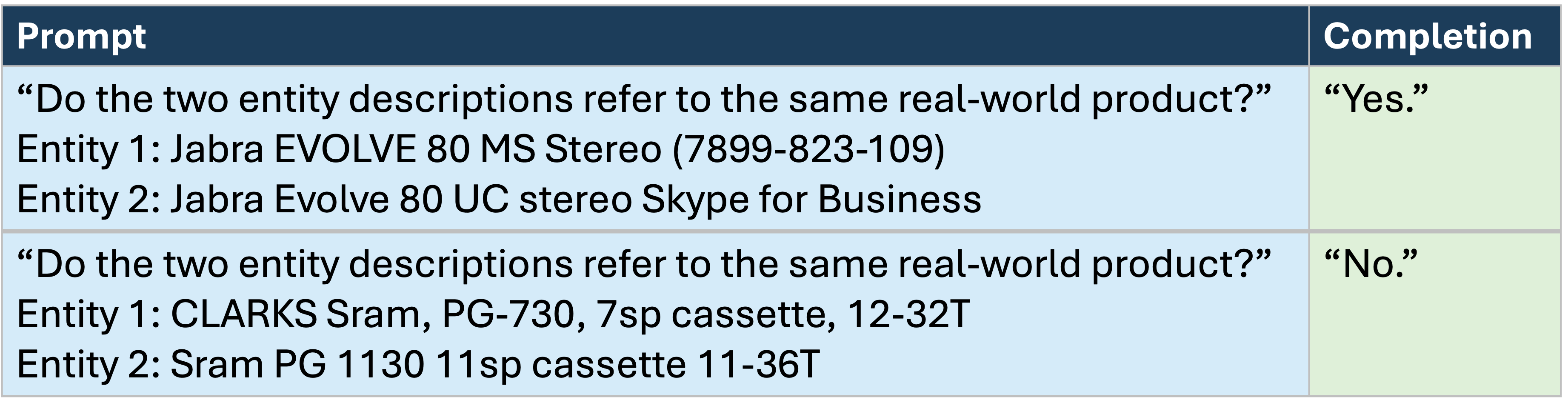}
\caption{Representation of training examples for standard fine-tuning.} 
\label{fig:standard_fine_tuning}
\end{figure}

To establish baselines, we report results from fine-tuning experiments using a simple, straightforward representation of training examples.
Figure \ref{fig:standard_fine_tuning} shows two examples from this set.
Each example consists of a prompt asking if the entity descriptions refer to the same entity, followed by both descriptions and the completion ``Yes'' or ``No'' based on whether they match.

Table~\ref{tab:standard_fine_tuning} reports results of standard fine-tuning experiments. Each row shows a specific LLM (column Model) fine-tuned on a given training set (column Training set). Subsequent columns display performance across multiple test sets. The \emph{zero-shot} rows show results for models without fine-tuning.


\begin{table*}[h!]
\centering
\caption{F1 scores after standard fine-tuning. The rows determine the model/training set combination, while the columns contain results for different test sets. The best result of each model per test set is set bold, the second best result is underlined.}
\resizebox{\textwidth}{!}{%
\begin{tabular}{llcccccccc}
\toprule
\multirow{2}{*}{Model} & \multirow{2}{*}{Training set} & \multicolumn{4}{c}{Product Domain} & Average & \multicolumn{2}{c}{Scholar Domain} & Average \\
\cmidrule(lr){3-6} \cmidrule(lr){8-9}
 &  & A-B & A-G & W-A & WDC & Gain & D-A & D-S & Gain\\
\midrule

\multirow{1}{*}{Llama 8B} & {Zero-shot} & 56.57 (0.00) & 49.16 (0.00) & 42.04 (0.00) & 53.36 (0.00) & - &  85.52 (0.00) & 67.69 (0.00) & - \\
\multirow{1}{*}{Llama 8B} & {A-B} & \textbf{87.34 (+30.77)} & \underline{59.16 (+10.00)} & \underline{60.39 (+18.35)} & \underline{66.07 (+12.71)} & 102\% & 79.60 (-5.92) & 42.89 (-24.80) & -83\% \\
\multirow{1}{*}{Llama 8B} & {A-G} & 67.48 (+10.91) & 50.00 (+0.84) & 44.73 (+2.69) & 39.53 (-13.83) & -1\% & 76.28 (-9.24) & 60.89 (-6.80) & -43\%  \\
\multirow{1}{*}{Llama 8B} & {W-A} & \underline{86.24 (+29.67)} & \textbf{60.41 (+11.25)} & \textbf{65.65 (+23.61)} & 57.80 (+4.44) & 96\% & 71.71 (-13.81) & 51.19 (-16.50) & -82\%  \\
\multirow{1}{*}{Llama 8B} & {WDC} & 81.78 (+25.21) & 52.29 (+3.13) & 53.74 (+11.70) & \textbf{69.19 (+15.83)} & 72\% & 74.52 (-11.00) & 67.40 (-0.30) & -30\%  \\
\multirow{1}{*}{Llama 8B} & {D-A} & 58.02 (+1.45) & 49.66 (+0.50) & 40.82 (-1.22) & 39.63 (-14.72) & -20\% & \textbf{97.42 (+11.90)} & \underline{79.56 (+11.87)} & 47\%  \\
\multirow{1}{*}{Llama 8B} & {D-S} & 65.71 (+9.15) & 46.22 (-2.95) & 42.35 (+0.31) & 52.00 (-1.35) & 7\% & \underline{96.70 (+11.18)} & \textbf{92.95 (+25.26)} & 94\%  \\

\midrule
\multirow{1}{*}{gpt-4o-m} & {Zero-shot} & 87.68 (0.00) & 59.20 (0.00) & 65.06 (0.00) & 81.61 (0.00) & - & 94.16 (0.00) & 87.96 (0.00) & - \\
\multirow{1}{*}{gpt-4o-m} & {A-B} & \textbf{94.09 (+6.41)} & 67.18 (+7.98) & 68.81 (+3.75) & \underline{82.69 (+1.08)} & 35\% & \underline{96.94 (+2.79)} & 88.85 (+0.89) & 28\% \\
\multirow{1}{*}{gpt-4o-m} & {A-G} & 83.51 (-4.17) & \textbf{80.25 (+21.05)} & \underline{68.97 (+3.91)} & 73.99 (-7.62) & -36\% & 96.28 (+2.12) & 85.60 (-2.37) & -2\% \\
\multirow{1}{*}{gpt-4o-m} & {W-A} & \underline{92.08 (+4.40)} & \underline{67.50 (+8.30)} & \textbf{78.85 (+13.79)} & 78.52 (-3.09) & 33\% & 95.58 (+1.43) & 86.97 (-0.99) & 3\% \\
\multirow{1}{*}{gpt-4o-m} & {WDC} & 91.44 (+3.76) & 64.11 (+4.91) & 68.92 (+3.86) & \textbf{84.38 (+2.77)} & 9\% & 85.35 (-8.81) & 76.33 (-11.63) & -155\% \\
\multirow{1}{*}{gpt-4o-m} & {D-A} & 88.94 (+1.26) & 67.32 (+8.12) & 67.51 (+2.45) & 81.34 (+0.27) & 27\% & \textbf{99.10 (+4.94)} & \underline{89.93 (+1.97)} & 24\% \\
\multirow{1}{*}{gpt-4o-m} & {D-S} & 89.76 (+2.08) & 65.71 (+6.50) & 68.46 (+3.39) & 70.87 (-10.73) & 3\% & {95.36 (+1.20)} & \textbf{96.22 (+8.26)} & 24\% \\

\midrule
\multirow{1}{*}{Llama 70B} & {Zero-shot}  & \textbf{79.12 (0.00)} & \underline{51.44 (0.00)} & \underline{55.62 (0.00)} & \textbf{75.19 (0.00)} & - & \textbf{80.50 (0.00)} & \textbf{69.47 (0.00)} & - \\
\multirow{1}{*}{Llama 70B} & WDC & \underline{77.94 (-1.18)} & \textbf{55.36 (+3.92)} & \textbf{60.56 (+4.94)} & \underline{72.66 (-2.53)} & - & \underline{69.90 (-10.60)} & \underline{63.85 (-5.62)} & - \\

\midrule
\multirow{1}{*}{gpt-4o} & {Zero-shot} & \textbf{92.20 (0.00)} & \underline{63.45 (0.00)} & \textbf{70.67 (0.00)} & \underline{81.64 (0.00)} & - & \underline{87.18 (0.00)} & \underline{74.59 (0.00)} & - \\
\multirow{1}{*}{gpt-4o} & WDC & \underline{91.99 (-0.21)} & \textbf{65.12 (+1.67)} & \underline{68.55 (-2.12)} & \textbf{87.07 (+5.43)} & - & \textbf{89.27 (+2.09)} & \textbf{80.74 (+6.15)} & - \\

\bottomrule
\end{tabular}
}
\label{tab:standard_fine_tuning}
\end{table*}

\subsection{Effectiveness}
This section discusses the results of testing the LLMs with examples from the same dataset (a non-transfer setting). The fine-tuned models show a performance increase on most datasets compared to the baseline models.

\textbf{Smaller Models:} 
Llama-8B shows a 17.31-point average F1 gain over zero-shot baseline (Table \ref{tab:standard_fine_tuning}, upper section). 5 of 6 models significantly improve, with Abt-Buy achieving the highest gain (30.77 points, Table \ref{tab:standard_fine_tuning}, A-B column). Amazon-Google is an outlier with only 0.84 points improvement (Table \ref{tab:standard_fine_tuning}, A-G column). GPT-4o-mini demonstrates an 11.72-point average increase (Table \ref{tab:standard_fine_tuning}, rows gpt-4o-m). Unlike Llama-8B, GPT-4o-mini achieves substantial gains on Amazon-Google, its highest absolute improvement (Table \ref{tab:standard_fine_tuning}, A-G column).

\textbf{Large Models:} 
Due to resource constraints, we fine-tune larger models only on the WDC small training set. Llama 70B experiences a performance decrease (-2.53 points, Table \ref{tab:standard_fine_tuning}, WDC column). In contrast, GPT-4o improves by 5.43 points to reach an F1 score of 87.07 on WDC, the highest result observed so far, despite starting with a similar zero-shot score as GPT-4o-mini.

\subsection{Generalization}
We evaluate two types of generalization: in-domain, meaning that generalization is assessed across datasets within the same topical domain, and cross-domain, where a model is trained using a training set from one domain and tested with a test set belonging to a dataset from a different topical domain (e.g., from products to scholar).
In order to measure how well a model that was fine-tuned using training data from dataset ($D_{\text{source}}$) generalizes to a target dataset ($D_{\text{target}}$), we compare the  performance of the model before and after fine-tuning (\emph{performance gain}) to the performance gain of a model that was fine-tuned directly for $D_{\text{target}}$ using the train set of $D_{\text{target}}$.
Let:
\begin{itemize}
    \item $F_0^{(t)}$ be the \emph{zero-shot} F1 score of a model on $D_{\text{target}}$,
    \item $F_{\text{transfer}}^{(t)}$ the F1 of a model \emph{fine-tuned on $D_{\text{source}}$} 
          but evaluated on $D_{\text{target}}$, and
    \item $F_{\text{target}}^{(t)}$ the F1 of a model \emph{fine-tuned and evaluated} on $D_{\text{target}}$,
\end{itemize}

For each target dataset $t$, we define the \emph{transfer gain}:
\begin{equation}
\label{eq:transfer_score}
    \text{Transfer gain}(t) \;=\;
    \frac{F_{\text{transfer}}^{(t)} \;-\; F_0^{(t)}}{F_{\text{target}}^{(t)} \;-\; F_0^{(t)}}
\end{equation}
The transfer gain measures which fraction of the performance gain that can be achieved by fine-tuning using training data from $D_{\text{target}}$ is achieved by fine-tuning using training data from $D_{\text{source}}$.
Since we have multiple target datasets from the same topical domain (e.g., product-related datasets like Abt-Buy and Walmart-Amazon), we report the average of all transfer gains from a source dataset $D_{\text{source}}$ to all target datasets ($T$) within a topical domain in the following.



\textbf{In-domain Generalization:}
In the product domain, 5/6 Llama 8B models improve performance across both their own and other datasets. Amazon-Google is the only model showing decline in a in-domain generalization setting (Table \ref{tab:standard_fine_tuning}, WDC column). On average, Llama 8B models achieve 59\% of the performance in an in-domain transfer scenario compared to models that were fine-tuned directly for the target datasets.

GPT-4o-mini exhibits less transfer capability, achieving on average a only 15\% performance increase relative to dataset-specific models, primarily due to poor Amazon-Google and Walmart-Amazon performance on WDC (Table \ref{tab:standard_fine_tuning}).

For larger models (fine-tuned only on WDC), Llama 70B generalizes successfully to Amazon-Google and Walmart-Amazon (+3.92 and +4.94 F1) but underperforms on Abt-Buy (-1.18). GPT-4o, despite higher WDC gains, generalizes poorly, outperforming baseline on only 1/3 datasets.

In the scholar domain, Llama 8B and GPT-4o-mini achieve better generalization (62\% and 66\% respectively), likely due to higher dataset similarity, as both benchmarks include DBLP records.

\textbf{Cross-Domain Generalization:}
In product-to-scholar transfer, Llama 8B models average a decrease of 11.05 F1 points, while GPT-4o-mini shows a 2.07-point decrease compared to zero-shot. All Llama 8B and half of GPT-4o-mini models underperform their zero-shot baselines (Table \ref{tab:standard_fine_tuning}, DBLP-ACM and DBLP-Scholar columns).

In scholar-to-product transfer, Llama 8B models underperform by 1.11 F1 points, while GPT-4o-mini gains 1.60 points on average. Larger models follow similar patterns: Llama 70B fails to exceed zero-shot in cross-domain scenarios, while GPT-4o gains 4.12 F1 points (Table \ref{tab:standard_fine_tuning}, scholar domain columns).

In conclusion, standard fine-tuning significantly enhances in-domain generalization for smaller models, while results for larger models and cross-domain generalization remain mixed. 

\section{Dimension 1: Example Representation}

This section investigates how different training example representations affect the performance and generalization capabilities of fine-tuned models. We compare three representations: 1) simple pairs as used in Section \ref{sec:standard_fine_tuning}, 2) pairs augmented by the LLM with textual explanations why the entities match or do not match, and 3) pairs augmented with structured explanations. The complete prompts for instructing the model to generate explanations as well as the augmented training sets are found in the project repository.

\begin{figure}[h]
\centering
\includegraphics[width=0.9\linewidth]{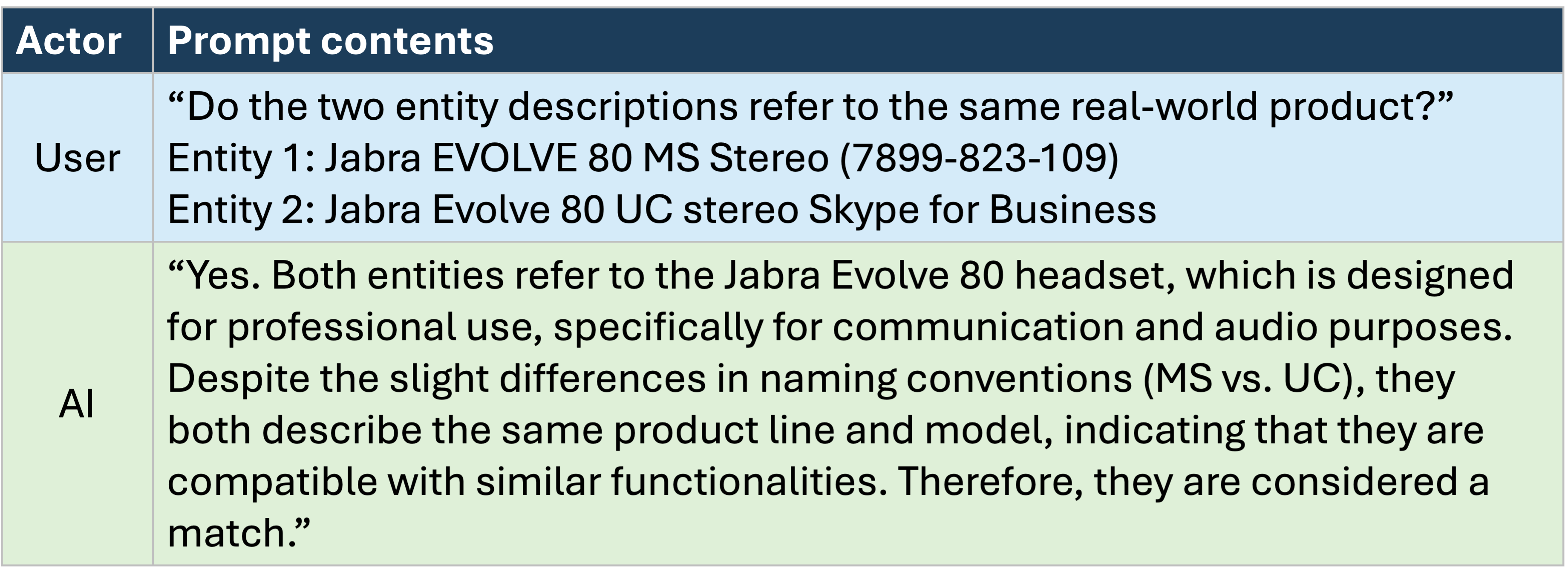}
\caption{Training example containing a textual explanation.} 
\label{fig:aws_explanation}
\end{figure}

\textbf{Textual Explanations:} 
We evaluate two types of textual explanations generated using GPT-4o-mini. The first approach presents the model with the prompt shown in Figure \ref{fig:standard_fine_tuning}, followed by the label. Afterwards, the model is asked to explain the label. The model is given no guidance on the format of the explanation. This results in detailed explanations, having an average length of 293 tokens, which describe why the entity descriptions match or do not match.
As a second approach, we re-run the technique proposed by Wadhwa et al.~\cite{wadhwa2024}, which generates more concise explanations by including short explanations for other entity pairs as demonstrations in the prompt. The resulting explanations have an average length of 90 tokens. Figure \ref{fig:aws_explanation} shows a training example containing an explanation in the AI part which was generated using this approach. The prompts used for generating these explanations, as well as the longer, more open-ended explanations, are available in the accompanying repository.

\begin{figure}[h]
\centering
\includegraphics[width=0.9\linewidth]{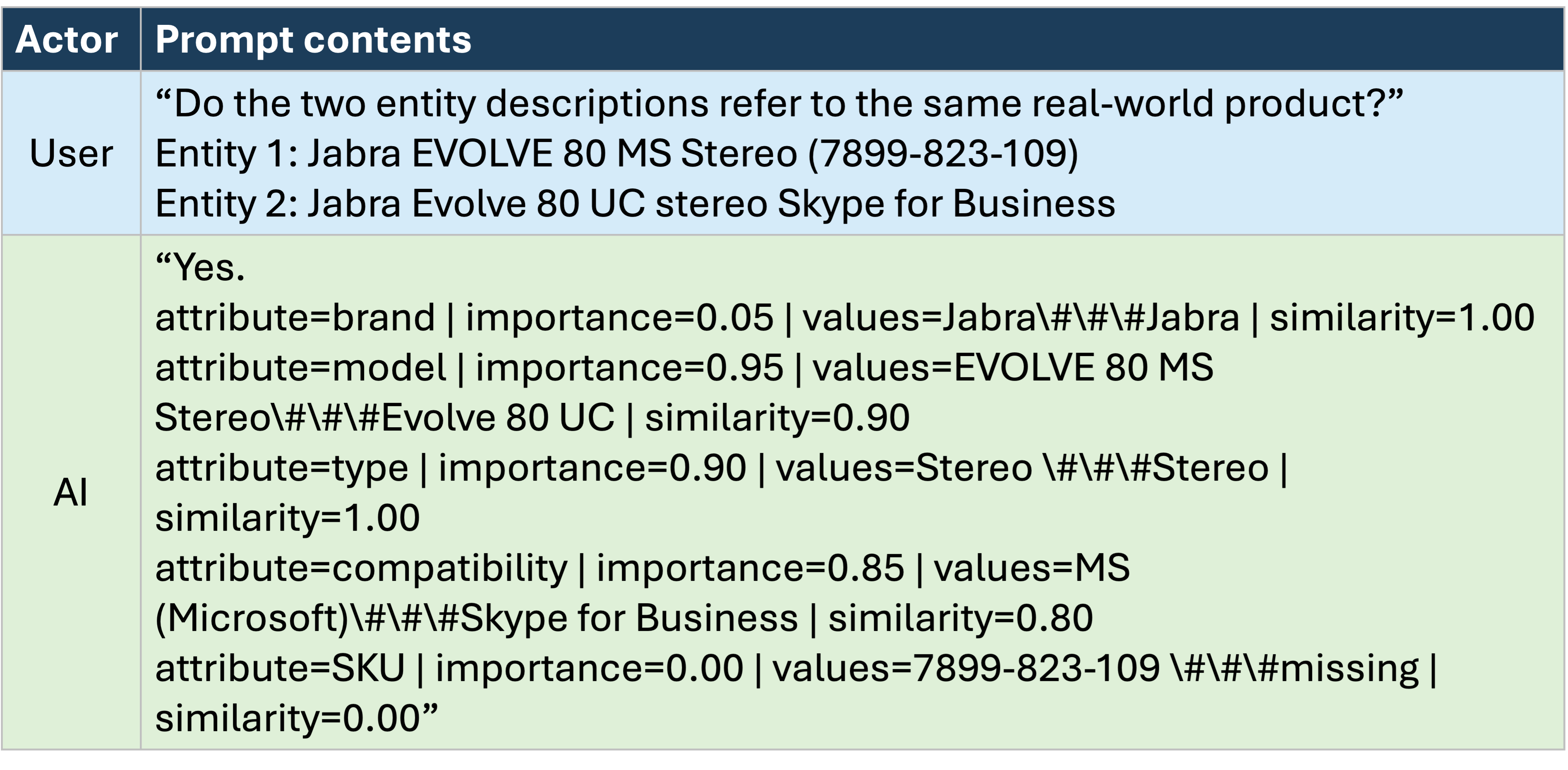}
\caption{Training example containing a Structured explanation.} 
\label{fig:structured_explanation}
\end{figure}

\textbf{Structured Explanations:} 
Our third approach is based on the assumption that LLMs can learn more from structured explanations compared to textual ones. Thus, building on the work of Peeters et al.~\cite{peeters2025entitymatchingusinglarge}, we instruct the model to generate explanations in a structured format that explicitly mention the attributes that were used for the matching decision, their importance for the decision, as well as the similarity of the attribute values. 
Figure \ref{fig:structured_explanation} shows an example of a structured explanation that was generated by GPT-4o-mini. The complete prompt for the generation of the explanation is found in the project repository and is discussed in ~\cite{peeters2025entitymatchingusinglarge}.
The example shows that the LLM incorporates factors beyond string matching for calculating similarities, such as recognizing the relationship between Microsoft and Skype for Business.

\begin{table*}[h!]
\centering
\caption{Fine-tuning results for different types of explanations (F1 scores, deltas to standard fine-tuning on WDC).}
\resizebox{\textwidth}{!}{%
\begin{tabular}{llcccccccc}
\toprule
\multirow{2}{*}{Model} & \multirow{2}{*}{Train set} & {No Transfer} & \multicolumn{3}{c}{In-Domain Transfer} &Average & \multicolumn{2}{c}{Cross-Domain Transfer} & Average \\
\cmidrule(lr){3-3} \cmidrule(lr){4-6} \cmidrule(lr){8-9}
 & & WDC & A-B & A-G & W-A & Gain & D-A & D-S & Gain \\
\midrule
\multirow{1}{*}{Llama 8B} & {Zero-shot} & 53.36 (-15.83) & 56.57 (-25.21) & 49.16 (-3.13) & 42.04 (-11.70) & - & \underline{85.52 (+11.00)} & \underline{67.69 (+0.30)} & - \\
\multirow{1}{*}{Llama 8B} & {WDC} & 69.19 (0.00) & 81.78 (0.00) & 52.29 (0.00) & 53.74 (0.00) & 72\% & 74.52 (0.00) & 67.40 (0.00) & -30\% \\

\multirow{1}{*}{Llama 8B} & {long textual} & 70.67 (+1.48) & 83.33 (+1.55) & 45.95 (-6.34) & 46.53 (-7.21) & 51\% & 51.11 (-23.41) & 47.92 (-19.48) & -146\% \\
\multirow{1}{*}{Llama 8B} & {Wadhwa et al.} & 73.20 (+4.01) & 79.00 (-2.78) & 50.30 (-1.99) & 48.90 (-4.84) & 55\% & 69.14 (+5.39) & 63.35 (-4.05) & -56\%\\

\multirow{1}{*}{Llama 8B} & {no imp.\&sim.} & 73.58 (+4.39) & \underline{85.25 (+3.47)} & \underline{52.56 (+0.27)} & 55.76 (+2.02) & 83\% & 55.55 (-18.98) & 51.14 (-16.25) & -125\% \\
\multirow{1}{*}{Llama 8B} & {no importance} & \underline{73.82 (+4.63)} & 84.82 (+3.04) & \textbf{54.26 (+1.97)} & \textbf{60.00 (+6.26)} & 93\% & \textbf{86.06 (+11.63)} & \textbf{69.19 (+1.79)} & 5\%\\
\multirow{1}{*}{Llama 8B} & {structured} & \textbf{74.13 (+4.94)} & \textbf{86.89 (+5.11)} & 51.84 (-0.45) & \underline{59.32 (+5.58)} & 91\% & 79.88 (+5.36) & 63.67 (-3.72) & -26\% \\

\midrule

\multirow{1}{*}{gpt-4o-m} & {Zero-shot} & 81.61 (-1.80) & 87.68 (-2.77) & 59.20 (-3.09) & 65.06 (-2.39) & - & \textbf{94.16 (+8.81)} & \underline{87.96 (+11.63)} & - \\
\multirow{1}{*}{gpt-4o-m} & {WDC} & \underline{83.41 (0.00)} & 90.45 (0.00) & \underline{62.29 (0.00)} & \underline{67.45 (0.00)} & 13\% & 85.35 (0.00) & 76.33 (0.00) & -55\% \\

\multirow{1}{*}{gpt-4o-m} & {long textual} & 81.30 (-2.11) & 88.94 (-1.51) & 61.37 (-0.92) & 64.23 (-3.22) & 5\% & 89.75 (+4.40) & \textbf{88.10 (+11.77)} & -11\% \\
\multirow{1}{*}{gpt-4o-m} & {Wadhwa et al.} & 80.81 (-2.60) & 84.12 (-6.33) & 59.03 (-3.26) & 64.19 (-3.26) & -14\% & \underline{93.18 (+7.83)} & 87.77 (+11.44) & -3\%  \\

\multirow{1}{*}{gpt-4o-m} & {no imp.\&sim.} & 81.04 (-2.37) & \underline{90.95 (+0.50)} & 61.30 (-0.99) & 66.40 (-1.05) & 7\% & 92.80 (+7.45) & 85.73 (+9.40) & -10\%  \\
\multirow{1}{*}{gpt-4o-m} & {no importance} & 83.17 (-0.24) & 90.26 (-0.19) & 60.71 (-1.58) & 65.09 (-2.36) & 4\% & 90.51 (+5.16) & 84.82 (+8.49)  & -18\% \\
\multirow{1}{*}{gpt-4o-m} & {structured} & \textbf{84.38 (+0.97)} & \textbf{91.44 (+0.99)} & \textbf{64.11 (+1.82)} & \textbf{68.92 (+1.47)} & 23\% & {88.87 (+3.52)} & {79.45 (+3.12)}  & -37\% \\

\midrule

\multirow{1}{*}{Llama 70B} & {Zero-shot} & \underline{75.20 (+2.50)} & \underline{79.10 (+1.20)} & 51.40 (-4.00) & 55.60 (-5.00) & - & \textbf{80.50 (+10.60)} & \textbf{69.50 (+5.60)}  & - \\
\multirow{1}{*}{Llama 70B} & {WDC} & 72.70 (0.00) & 77.90 (0.00) & \textbf{55.40 (0.00)} & \underline{60.60 (0.00)} & - & 69.90 (0.00) & \underline{63.90 (0.00)}  & - \\
\multirow{1}{*}{Llama 70B} & {structured} & \textbf{76.70 (+4.00)} & \textbf{84.80 (+6.90)} & \underline{52.80 (-2.60)} & \textbf{65.80 (+5.20)} & - & \underline{70.10 (+0.20)} & 62.10 (-1.80)  & - \\

\midrule

\multirow{1}{*}{gpt-4o} & {Zero-shot} & 81.60 (-5.50) & \textbf{92.20 (+0.20)} & \underline{63.45 (-1.65)} & \textbf{70.67 (+2.17)} & - & \underline{87.18 (-2.09)} & {74.59 (-6.15)}  & - \\
\multirow{1}{*}{gpt-4o} & {WDC} & \textbf{87.10 (0.00)} & \underline{92.00 (0.00)} & \textbf{65.10 (0.00)} & \underline{68.50 (0.00)} & - & \textbf{89.27 (0.00)} & \textbf{80.74 (0.00)}  & - \\
\multirow{1}{*}{gpt-4o} & {structured} & \underline{83.20 (-3.90)} & 90.60 (-1.40) & 62.80 (-2.30) & 66.50 (-2.00) & - & 84.69 (-4.58) & \underline{74.90 (-5.84)}  & - \\

\bottomrule
\end{tabular}
}
\label{tab:explanation_fine_tuning}
\end{table*}

\subsection{Effectiveness}
We compare the effectiveness of different explanation approaches in a non-transfer scenario using the WDC Products dataset. The result of these experiments are found in the No Transfer column of Table \ref{tab:explanation_fine_tuning}.

\textbf{Smaller Models:}
Llama 8B outperforms standard fine-tuning across all evaluated explanation approaches, as shown in Table \ref{tab:explanation_fine_tuning}, column No Transfer. Long textual explanations provide the smallest increase of 1.48 F1 points. 
The approach by Wadhwa et al. results in a 4.01-point improvement. Structured explanations further improve F1 scores by 0.93 over Wadhwa’s approach.
GPT-4o-mini shows a different trend, where only structured explanations outperform the standard fine-tuning, increasing by 0.97 F1 points (Table \ref{tab:explanation_fine_tuning}, WDC Products column). The other configurations underperform, with three out of five approaches performing worse than zero-shot (Table \ref{tab:explanation_fine_tuning}, Non-transfer column).

\textbf{Larger Model:}
Structured explanations are exclusively tested on larger models due to their consistent performance improvement in smaller models. While Llama 70B does not benefit from standard fine-tuning, structured explanations improve its performance by 1.50 F1 points over zero-shot (Table \ref{tab:explanation_fine_tuning}, WDC column).
In contrast, GPT-4o does not follow this trend. While structured explanations lead to a 1.60-point increase over zero-shot, standard fine-tuning results in a 5.50-point improvement, as shown in the GPT-4o rows in the WDC column of Table \ref{tab:explanation_fine_tuning}. Therefore, GPT-4o does not significantly benefit from structured explanations in a non-transfer setting.

\subsection{Generalization}
We fine-tune the models using the WDC dataset and afterwards apply them to the test sets of the other datasets. We measure the generalization performance within the product domain and across domains using the scholarly datasets.

\textbf{In-Domain Generalization:}
Standard fine-tuning results in Llama 8B achieving 72\% of the dedicated models’ performance on product datasets (Table \ref{tab:explanation_fine_tuning}, In-Domain Average Gain column). The Wadhwa et al. approach and longer textual explanations underperform, achieving 55\% and 51\%, respectively. However, structured explanations surpass the fine-tuned baseline, with the \textit{structured} and \textit{no importance} approaches achieving 93\% and 91\% of source model performance (Table \ref{tab:explanation_fine_tuning}, In-Domain Average Gain column).
GPT-4o-mini continues the trend where 4 out of 5 explanation approaches underperform the fine-tuned baseline. However, structured explanations outperform the baseline, achieving 23\% of the source models’ performance compared to 13\% by the fine-tuned baseline (Table \ref{tab:explanation_fine_tuning}, In-Domain Average Gain column).
Llama 70B benefits from structured explanations, outperforming both the fine-tuned baseline and zero-shot in in-domain generalization. It shows the highest generalization performance on two out of three product datasets (Table \ref{tab:explanation_fine_tuning}, Llama 70B structured columns A-B and W-A). This trend however does not extend to GPT-4o as structured explanations underperform in this instance.
Llama 70B fine-tuned with structured explanations outperforms the other approaches. Specifically, it achieves the highest generalization gain on two out of three product datasets (Table \ref{tab:explanation_fine_tuning}, Llama 70B structured, columns A-B and W-A). However, this improvement does not extend to GPT-4o, where structured explanations negatively impact generalization performance.

\textbf{Cross-Domain Generalization:}
Llama 8B struggles to generalize effectively in cross-domain settings, with all approaches underperforming compared to zero-shot performance. In contrast, GPT-4o-mini shows improvements across all explanation approaches, however non-rival zero-shot performance.
In conclusion, structured explanations significantly enhance in-domain generalization,  while cross-domain generalization remains limited.

\section{Dimension 2: Example Selection and Generation}

This section explores how the selection and generation of training examples for fine-tuning impact in-domain performance and model generalization. We examine three strategies: filtering the existing training sets, generating new artificial examples based on existing data, and adding additional training examples that are similar to examples on which the model failed. We evaluate the techniques without augmenting the training examples with explanations. We provide the complete prompts for the example selection and generation as well as filtered and augmented training sets in the project repository.

\subsection{Training Set Filtration}
The filtration approach is based on the assumption that training data quality outweighs quantity and it is thus beneficial to remove potentially misleading examples from the training set. This is in accordance with the documentation of OpenAI who recommend smaller training sizes in favor of presenting examples of missbehaviour by the model\footnote{https://platform.openai.com/docs/guides/fine-tuning}.

\textbf{Error-based Filtering:} This approach attempts to filter out mislabeled training examples from WDC training set. For this purpose, we use GPT-4o-mini and discard all examples that are incorrectly labeled by the model. This approach may discard some correctly labeled examples. However, it is hypothesized that overall dataset quality improves. The resulting training set WDC-filtered has a size of 2006 examples, compared to 2500 examples of the original WDC-small training set (see Table \ref{tab:filtration}).

\textbf{Relevancy-based Filtering:}  The second approach involves filtering for relevance, operating on the assumption that fewer, more meaningful examples lead to better outcomes. For example, comparing a hard drive and a TV might pass initial filtration but offer limited value due to their low complexity. By removing such examples, computational demands are certainly reduced, while training results may potentially be enhanced. This approach prompts GPT-4o to select only ``interesting'' training examples, a term we intentionally leave undefined, allowing the model to interpret it. The model appears to define ``interesting'' as pairs that share many attributes and are thus highly similar (corner cases). Pairs deemed irrelevant by the model are discarded. The resulting WDC-filtered-rel training set has a total size of 608 examples (see Table \ref{tab:filtration}).

\textbf{Effectiveness:}
As shown in Table \ref{tab:selection_generation_fine_tuning} column No Transfer, the Llama 8B model trained on the \textit{WDC-s-filtered} and \textit{WDC-s-filtered-rel} datasets outperforms the fine-tuned baseline by 4.73 and 3.18 F1 points, respectively. Both filtered approaches surpass the Llama 8B model trained on the WDC large training set, containing nearly 20,000 examples (see Table~\ref{tab:dataset_stats}). In contrast, GPT-4o-mini does not exhibit similar improvements, as both filtered datasets perform below the zero-shot model.

\textbf{Generalization:}
While filtration improves Llama 8B’s performance, it does not significantly impact in-domain generalization. The fine-tuned baseline captures 72\% of dedicated performance, while filtered approaches capture 75\% and 76\% with relevancy filtration (see Table \ref{tab:selection_generation_fine_tuning}, In-Domain Average Gain column). Since GPT-4o-mini does not improve its performance with filtration in a no-transfer scenario, the same trend is observed in in-domain generalization, where this approach underperforms the fine-tuned baseline across all product datasets.

For cross-domain generalization, both Llama 8B and GPT-4o-mini continue to underperform relative to the zero-shot baselines. 

\begin{table}[t]
\centering
\caption{Training set sizes after filtering and generation steps.}
\resizebox{0.75\columnwidth}{!}{%
\begin{tabular}{lccc}
\toprule
\textbf{Dataset} & \textbf{\# Pos} & \textbf{\# Neg} & \textbf{\# Total} \\
\midrule
WDC-small & 500 & 2000 & 2500 \\
WDC-filtered & 445 & 1561 & 2006 \\
WDC-filtered-rel & 442 & 166 & 608 \\
Syn & 4932 & 15208 & 20140 \\
Syn-filtered & 3264 & 10560 & 13824 \\
Syn-filtered-rel & 2182 & 6718 & 8900 \\
\bottomrule
\end{tabular}
}
\label{tab:filtration}
\end{table}

\begin{table*}[h!]
\centering
\caption{Fine-tuning with example selection and generation. Deltas against fine-tuning on WDC-small.}
\resizebox{\textwidth}{!}{%
\begin{tabular}{llcccccccc}
\toprule
\multirow{2}{*}{Model} & \multirow{2}{*}{Train set} & {No Transfer} & \multicolumn{3}{c}{In-Domain Transfer} & Average & \multicolumn{2}{c}{Cross-Domain Transfer} & Average \\
\cmidrule(lr){3-3} \cmidrule(lr){4-6} \cmidrule(lr){8-9}
 & & WDC & A-B & A-G & W-A & Gain & D-A & D-S & Gain \\
\midrule
\text{Llama 8B} & \text{Zero-shot} & 53.36 (-15.83) & 56.57 (-25.21) & 49.16 (-3.13) & 42.04 (-11.70) & - & \textbf{85.52 (+11.00)} & \underline{67.69 (+0.30)} & - \\
\text{Llama 8B} & \text{WDC-small} & 69.19 (0.00) & 81.78 (0.00) & 52.29 (0.00) & 53.74 (0.00) & 72\% & 74.52 (0.00) & 67.40 (0.00) & -30\% \\
\text{Llama 8B} & \text{WDC-medium} & 67.45 (-1.74) & 78.80 (-2.98) & 52.93 (+0.64) & 54.89 (+1.15) & 70\% & 75.06 (+0.54) & {65.22 (-2.18)} & -35\% \\
\text{Llama 8B} & \text{WDC-large} & 72.13 (+2.94) & 70.06 (-11.72) & 44.89 (-7.40) & 48.50 (-5.24) & 28\% & 78.47 (+3.94) & 56.95 (-10.44) & -48\% \\
\text{Llama 8B} & \text{WDC-s-filter} & 73.92 (+4.73) & {85.12 (+3.34)} & 49.47 (-2.82) & 54.51 (+0.77) & 75\% & \underline{80.89 (+6.37)} & \textbf{74.29 (+6.89)} & 5\% \\
\text{Llama 8B} & \text{WDC-s-filter-rel} & 72.37 (+3.18) & 79.43 (-2.35) & \textbf{54.73 (+2.44)} & 55.68 (+1.94) & 76\% & 76.49 (+1.97) & 66.11 (-1.29) & -29\% \\
\text{Llama 8B} & \text{Syn-filter} & 72.54 (+3.35) & 80.98 (-0.80) & 51.25 (-1.04) & \underline{56.65 (+2.91)} & 74\% & 68.37 (-6.15) & 57.23 (-10.17) & -74\% \\
\text{Llama 8B} & \text{Syn-filter-rel} & \underline{74.04 (+4.85)} & \textbf{86.00 (+4.22)} & \textbf{54.73 (+2.44)} & \textbf{59.48 (+5.74)} & 97\% & {75.06 (+0.53)} & {67.20 (-0.20)} & -29\% \\
\text{Llama 8B} & \text{WDC-s-err-sel} & \textbf{74.37 (+5.18)} & \underline{85.19 (+3.41)} & 52.88 (+0.59) & 55.80 (+2.06) & 83\% & 61.99 (-12.53) & 55.32 (-12.08) & -97\% \\

\midrule

\multirow{1}{*}{gpt-4o-m} & {Zero-shot} & \underline{77.44 (-5.87)} & \underline{85.47 (-4.78)} & 57.20 (-5.14) & \textbf{64.03 (+1.61)} & - & \underline{94.16 (+8.81)} & \textbf{87.96 (+11.63)} & - \\
\multirow{1}{*}{gpt-4o-m} & {WDC-small} & \textbf{83.31 (0.00)} & \textbf{90.25 (0.00)} & \textbf{62.34 (0.00)} & \underline{62.42 (0.00)} & 9\% & 75.65 (0.00) & 76.33 (0.00) & -55\% \\
\multirow{1}{*}{gpt-4o-m} & {WDC-s-filter} & 77.06 (-6.25) & 81.38 (-8.87) & 44.67 (-17.67) & 49.84 (-12.58)  & -61\% & 92.89 (+7.54) & 78.34 (+2.01) & -29\% \\
\multirow{1}{*}{gpt-4o-m} & {Syn-filter} & 76.89 (-6.42) & 84.84 (-5.41) & \underline{60.29 (-2.05)} & 61.67 (-0.75)  & -2\% & \textbf{94.84 (+9.49)} & \underline{79.32 (+2.99)} & -21\% \\

\bottomrule
\end{tabular}
}
\label{tab:selection_generation_fine_tuning}
\end{table*}

\subsection{Example Generation}
While the earlier approaches focused on discarding irrelevant examples, this section investigates using GPT-4o to generate additional training examples for fine-tuning. 
We test three methods for generating examples. All methods iterate over the WDC small training set and use the examples from this set as seeds to derive additional training examples. The complete prompts used by the methods are provided in the project repository.

\textbf{Brief Generation Prompt:} This method provides the model with a short task description, instructing it to generate three non-matches and one match. The prompt also includes a seed pair from the training set.

\textbf{Detailed Generation Prompt:} This prompt contains a comprehensive task description, including background information on entity matching and specific requirements for the generated examples. It instructs the model to produce examples from the same product category and to include similar matching challenges as the challenges found in the provided seed example. In addition, the prompt explains key concepts such as corner cases to deepen the model’s understanding of the task.

\textbf{Demonstration-based Creation:} Building on the second method, this approach includes six entity pairs as demonstrations in the prompt. These pairs are selected from the WDC small dataset based on their similarity to the seed example in the OpenAI embedding space.

\textbf{Inspection of the Generated Examples:} We manually inspect a subset of the generated pairs. This inspection reveals that the first method encounters two main issues: generating matching examples with differing strings and maintaining correctness. The model often produces matching examples that are easy non-matches, such as laptops from different product lines being incorrectly treated as matches. The detailed generation prompt results in examples with more variation, though correctness remains mixed.
The third approach increases variance further. However, many of these examples remain inaccurate upon human evaluation.

\textbf{Combining Generation and Filtration:} Given the mixed quality of the generated examples, we combine example generation with the filtering approach introduced above, further refined by relevancy filtering (syn.-filter-rel.). The resulting dataset sizes are presented in Table \ref{tab:filtration}. Since the generation methods are intended to extend existing datasets, both training splits are combined with the unfiltered version of the WDC small dataset (Table \ref{tab:filtration}).

\textbf{Effectiveness:}
Table \ref{tab:selection_generation_fine_tuning} column No Transfer shows that the Llama 8B model gains 3.35 F1 points without relevancy filtration and 4.85 F1 points with relevancy filtration.
In contrast, GPT-4o-mini shows a 6.42 F1 point decrease compared to the fine-tuned baseline, underperforming zero-shot. Due to the high cost of fine-tuning the GPT series and the underperformance of this approach, we chose not to pursue fine-tuning with relevancy filtration.

\textbf{Generalization:}
In a non-transfer scenario, the generated examples perform similarly to the filtration approach. However, the additional examples enable the models to outperform other approaches in a generalization context. The generated examples with relevancy filtration capture 97\% of the dedicated model performance in an in-domain generalization scenario, compared to 72\% for the fine-tuned baseline.
For GPT-4o-mini, both trailed approaches fail to outperform the fine-tuned baseline in in-domain generalization performance.
Cross-domain transfer continues the trend of Llama underperforming in this regard, while GPT-4o-mini shows a 23\% improvement with generated examples.

\subsection{Error-based Example Selection}
The idea of the error-based example selection (err.-sel.) approach is to guide the model to prevent errors by fine-tuning it with examples that are similar to entity pairs that are matched incorrectly by the model. For this, the Llama 8B model is initially trained on the standard 2,500 examples from the WDC small training dataset. After this initial training phase, the model is validated to identify the remaining errors. Additional training examples are then selected from the large WDC products dataset, as shown in Table \ref{tab:dataset_stats}, simulating additional labeling capacity. These examples are chosen based on their embedding similarity to the identified errors.

Each iteration starts with 2,500 examples from the WDC small dataset, adding 2,500 more based on errors. The model is retrained for 5 epochs, repeating this process five times. The best model is selected based on the highest F1 score on the validation set. Due to OpenAI’s fine-tuning limitations and high computational cost, this approach is tested only with Llama 8B.

\textbf{Effectiveness:}
The ``WDC-s-err-sel'' approach yields the highest F1 score for Llama 8B among all tested methods, with a 5.18-point increase to 74.37, a 1.99-point improvement over the large WDC dataset. This result demonstrates that adding only highly relevant examples outperforms merely increasing training data quantity.

\textbf{Generalization:}
While this approach yields improved dedicated performance, generalization does not improve beyond that of example generation (83\% average gain).
Interestingly, this approach results in severe underperformance in a cross-domain transfer scenario, with the worst results out of all tested approaches.

\section{Cost Analysis}
\label{sec:cost_analysis}

This chapter presents a cost analysis for hosted LLMs. 
Table~\ref{tab:cost_comparison_gpt} lists the costs and the number of tokens used by GPT-4o-mini and GPT-4o for the WDC Products dataset across three scenarios: \textit{zero-shot}, \textit{standard} fine-tuning, and \textit{structured explanation} fine-tuning.

\begin{table}[ht]
\centering
\caption{Cost comparison GPT-4o-mini/4o on WDC-small.}
\resizebox{0.48\textwidth}{!}{%
\begin{tabular}{lcccccc}
\hline
 & \multicolumn{2}{c}{\textbf{Zero-shot}} 
 & \multicolumn{2}{c}{\textbf{Fine-tuned Standard}} 
 & \multicolumn{2}{c}{\textbf{Fine-tuned Structured}} \\
\cline{2-3}\cline{4-5}\cline{6-7}
\textbf{Metric} & \textbf{4o-mini} & \textbf{4o} & \textbf{4o-mini} & \textbf{4o} & \textbf{4o-mini} & \textbf{4o} \\
\hline
F1 score & 81.61 & 81.60 & 83.41 & 87.10 & 84.38 & 83.20 \\
\hline
\multicolumn{7}{l}{\textbf{Fine-tuning (2,500 examples)}} \\
Training tokens & 0 & 0 & 1,841,460 & 1,841,460 & 5,750,330 & 5,750,330 \\
Cost per example & 0 & 0 & 0.22¢ & 1.84¢ & 0.69¢ & 5.75¢ \\
Total fine-tuning cost & 0 & 0 & \$5.52 & \$46.04 & \$17.25 & \$143.76 \\
\hline
\multicolumn{7}{l}{\textbf{Inference (4,500 examples)}} \\
Total Input tokens & 338,735 & 338,735 & 338,735 & 338,735 & 338,735 & 338,735 \\
Total Output tokens & 4,500 & 4,626 & 4,500 & 4,500 & 14,758 & 14,683 \\
Mean Token Count & 76.27 & 76.30 & 76.27 & 76.27 & 78.55 & 78.54 \\
Total Token Count & 343,235 & 343,361 & 343,235 & 343,235 & 353,493 & 353,418 \\
Total Inference Cost & \$0.05 & \$0.89 & \$0.11 & \$1.34 & \$0.12 & \$1.49 \\
\hline
\end{tabular}%
}
\label{tab:cost_comparison_gpt}
\end{table}


At the time of our experiments, GPT-4o-mini costs \$0.15 (GPT-4o: \$2.50) per million \textit{input} tokens and \$0.60 (GPT-4o: \$10.00) per million \textit{output} tokens in a zero-shot setting. Fine-tuning is billed at \$3 (GPT-4o: \$25.00) per million \textit{training} tokens. Additionally, \textit{inference} on the resulting fine-tuned model doubles the token price to \$0.30 (GPT-4o: \$3.75) per million input tokens and \$1.20 (GPT-4o: \$15.00) per million output tokens.\footnote{\url{https://openai.com/api/pricing/} (as of January 2025)} This pricing structure generally makes fine-tuning more expensive than zero-shot usage.
As shown in Table~\ref{tab:cost_comparison_gpt}, zero-shot inference on GPT-4o-mini and GPT-4o yields comparable F1 scores (81.61 vs.\ 81.60), but GPT-4o is significantly more expensive at \$0.89 for inference compared to \$0.05 for GPT-4o-mini. Standard fine-tuning improves GPT-4o-mini’s F1 to 83.41 at a \$5.52 fine-tuning cost, while GPT-4o jumps to 87.10 at a fine-tuning cost of \$46.04. Structured fine-tuning increases GPT-4o-mini’s performance further (F1=84.38) for \$17.25, whereas GPT-4o drops slightly (F1=83.20) for \$143.76 in fine-tuning costs. Thus, while GPT-4o obtains a larger gain in standard fine-tuning, it also incurs substantially higher training and inference expenses. By contrast, GPT-4o-mini 
demonstrates that more elaborate techniques (e.g., structured fine-tuning) can remain cost-effective. Whether GPT-4o’s higher scores justify the tenfold cost depends on requirements of the specific use case. Ultimately, these findings highlight that a model’s raw performance gains must be weighed against sharply escalating token and fine-tuning costs, particularly for hosted large-scale models. The cost advantage of models without fine-tuning likely holds only in a hosted scenario; in self-hosted settings, there should not be a difference in the inference cost between a model without fine-tuning and with fine-tuning.

\section{Related work}
\label{sec:related_work}
Entity matching~\cite{Christen2012DataMC,elmagarmidDuplicateRecordDetection2007} has been studied extensively over the last 50 years~\cite{fellegiTheoryRecordLinkage1969}. Early methods relied on manually defined matching rules, while the field is now dominated by supervised machine learning techniques~\cite{christophides_end--end_2020, BarlaugNeural2021,liDeepEntityMatching2020,peetersDualobjectiveFinetuningBERT2021,yaoEntityResolutionHierarchical2022,zeakis2023pre,wangMachampGeneralizedEntity2021}. A prominent system leveraging pre-trained language models (PLMs), such as BERT, is \textsc{Ditto}\cite{liDeepEntityMatching2020}. PLM-based approaches typically require fewer computational resources, due to their smaller size, compared to LLM-based matchers~\cite{peeters2025entitymatchingusinglarge}. 

\textbf{Entity Matching using LLMs:} Narayan et al.\cite{narayan2022can} were the first to experiment with employing LLMs for entity matching. Subsequent studies\cite{peetersUsingChatGPTEntity2023a,fanCostEffectiveInContextLearning2023,nananukul2024costefficient} expanded this approach by evaluating further zero-shot and few-shot prompts on more recent models. Peeters, et al.~\cite{peeters2025entitymatchingusinglarge} investigate the effect of standard fine-tuning on the effectiveness and generalization performance of GPT4o-mini, Llama2-70B, and Llama3.1-70B. We build upon this work and experiment with varying the representation of training examples and use different example selection and generation methods. The fine-tuning results from ~\cite{peeters2025entitymatchingusinglarge} for specific datasets are not directly comparable to the results that we present in Chapter~\ref{sec:standard_fine_tuning}, as they subsample the test sets in order to save resources, while we use the complete test sets. 

\textbf{Fine-tuning using Explanations:} Zhang et al.\cite{zhangJellyfishLargeLanguage2023} explored fine-tuning LLMs via multi-task learning for multiple data integration tasks, including entity matching. They provide the models with reasoning traces in the training data, going beyond standard fine-tuning. Wadhwa et al.\cite{wadhwa2024} expanded upon this by incorporating textual explanations into the training process while focusing solely on entity matching. In contrast, we include structured explanations in the training data, hypothesizing that the consistent structure makes it easier for the model to learn from these explanations compared to purely textual approaches, such as~\cite{zhangJellyfishLargeLanguage2023, wadhwa2024}.

\textbf{Generalization Performance:} While earlier work~\cite{peeters2023wdc, wangEntityMatching2022} showed that PLMs, such as BERT, exhibit poor generalization for entity matching, \textsc{AnyMatch} by Zhang et al.\cite{zhang_anymatch_2024} demonstrates that smaller models can generalize in a leave-one-dataset-out evaluation setup. They also show that using attribute-level transformations for training data augmentation can increase the generalization performance. Notably, their use of GPT-2, a decoder-based architecture, differs from the encoder-based architectures studied in\cite{peeters2023wdc}. In contrast, our method explicitly generates novel synthetic examples, exploring a different strategy to further improve model robustness and generalization. In addition, we differentiate between in-domain and cross-domain generalization, taking into account the varying levels of domain shift.
Li, et al.~\cite{SIGMOD-MultiTaskLearning} investigate the ability of LLMs to generalize to unseen table tasks in a multi-task fine-tuning setup. 

\section{Conclusion}
\label{sec:conclusion}
This paper shows that by fine-tuning small LLMs for entity matching, it is possible to reach a performance comparable to that of large models. This significantly reduces both costs and the CO2 footprint associated with using LLMs for entity matching. Fine-tuning does not harm the model's ability to generalize to unseen datasets, while fine-tuning with structured explanations even improves this capability.
In future work, we plan to refine the example selection and generation methods and explore strategies to improve cross-domain generalization.

\balance
\bibliographystyle{IEEEtran}
\bibliography{sources.bib}

\end{document}